\documentclass[conference]{IEEEtran}
\usepackage{cite}
\usepackage{amsmath,amssymb,amsfonts}
\usepackage{algorithmic}
\usepackage{graphicx}
\usepackage{textcomp}
\usepackage{xcolor}
\def\BibTeX{{\rm B\kern-.05em{\sc i\kern-.025em b}\kern-.08em
    T\kern-.1667em\lower.7ex\hbox{E}\kern-.125emX}}
\begin{document}

\title{IMPROVEMENT IN SEMANTIC ADDRESS MATCHING USING NATURAL LANGUAGE PROCESSING \\
\thanks{}
}
\author{\IEEEauthorblockN{Vansh Gupta\IEEEauthorrefmark{1},
Mohit Gupta\IEEEauthorrefmark{2}, Jai Garg\IEEEauthorrefmark{3} and
Nitesh Garg\IEEEauthorrefmark{4}}
\IEEEauthorblockA{Electronics and Communication Engineering Department,\\
Dr. Akhilesh Das Gupta Institute of Technology \& Management, \\
New Delhi, India\\
Email: \IEEEauthorrefmark{1}vgvansh25@gmail.com,
\IEEEauthorrefmark{2}guptamohit1504@gmail.com,
\IEEEauthorrefmark{3}jaigarg2@gmail.com,
\IEEEauthorrefmark{4}nitesh.garg049@gmail.com}}

\maketitle

\begin{abstract}
Address matching is an important task for many businesses especially delivery and take out companies which help them to take out a certain address from their data warehouse. Existing solution uses similarity of strings, and edit distance algorithms to find out the similar addresses from the address database, but these algorithms could not work effectively with redundant, unstructured, or incomplete address data. This paper discuss semantic Address matching technique, by which we can find out a particular address from a list of possible addresses. We have also reviewed existing practices and their shortcoming. Semantic address matching is an essentially  NLP task in the field of deep learning. Through this technique We have the ability to triumph the drawbacks of existing methods like redundant or abbreviated data problems. The solution uses the OCR on invoices to extract the address and create the data pool of addresses. Then this data is fed to the algorithm BM-25 for scoring the best matching entries. Then to observe the best result, this will pass through BERT for giving the best possible result from the similar queries. Our investigation exhibits that our methodology enormously improves both accuracy and review of cutting-edge technology existing techniques.
\end{abstract}


\section{Introduction}
With increase in resources that are dependent on position, the need of address matching is also increasing. The companies has a lot of data of address and know whether they have this address in their database or not. Address matching is a complex task, as there are number of possible formats of writing an address. Moreover spelling mistakes and use of abbreviations also makes this task more hectic and difficult. 

For efficient address traversal and interpretation, address matching task has come into picture [1], focusing at recognizing addresses pointing to similar different place across various data on the address pools. Existing address tuning and normalisation methods depend on string matching measures such as edit distance, TF-IDF is not the case so good to tackle various forms of address expressions [2]. Constructing a conclusion tree out of acquired knowledge of matching rules is a more widely used mechanism for address matching, in which each law correlate with the tree, there is a path from the beginning to the leaf [3].

Some work merge to measure distance, edit distance with a space vector model. the resemblance of the strings. They estimate the variation caused by making use of edit-distance to distinguish between characters in an address, and compute the heterogeneity caused by TF-IDF term weighting differences in address. Weighting the two dissimilarities yields the final result [4]. To address the loopholes of the previous methods, Throughout this paper, rather than relying solely on the syntactic features of addresses, we assess the semantic similarity of addresses. with the assistance of NLP. Currently, deep learning techniques and NLP have made incredible progress on sentence description, such as Seq2Seq model for the purpose of computer translation [5] and the skip-thought model for distributed sentence representation [6].

The suggested model in this article is carry through following steps. The BM25 The model is employed first. as it considers the period of the log and the frequency of the term query word and score each most similar output query. Second, the enhanced BERT algorithm to find out the best similar result by passing the entries which have less than 3 units of difference in their score. By bringing the profound learning engineering into address coordinating, By developing this paper, it fills a gap by a viable and precise strategy for address coordinating with that  thinks about the strict similitude between address records yet in addition underlines their semantic connection dependent regarding 'address comprehension': the proposed technique can get high prescient precision whether the analyzed location records are indistinguishable, or just have not many strict covers [7]. 

Subsequently, our strategy can beat the current techniques for address coordinating in any event, when the address in question is unstructured or has a complicated syntax.

\section{RELATED WORK}
The String-based method, Dictionary-based method, and Address Matching Tree method are examples of existing methods. The String-Based approach incorporates the space of vectors display and changes the distance to assess position comparability [8].

The Dictionary-Based technique traverses the location components using a position  guide in order to reduce the effect of ambiguous spot names on address matching [9-10].

The Address Matching Tree (AMT) technique By converting the location coordinating with tree into a set of coordinating with guidelines, establishes a standard-based coordinating with a tree and applying it to coordinating with a tree [11].

Edit Distance-Based Text Matching Algorithm [12] Minimum Edit Distance (MED) calculation is utilized to figure the likeness degree between the information base of addresses and the information question, which we use to compute coordinating with degree between the two. Likewise the calculation is applied in the position looking through framework [13].

\section{PROPOSED SYSTEM}
\subsection{\textbf{Dataset Collection}}\label{AA}
Our experiment was based on addresses from different countries including USA, India, and Canada etc. To create a Database of address we collected, invoices of different companies, available of the internet using Web scraping. After collecting invoices, we applied OCR using tesseract to extract the different addresses. A sample of addresses have been shown below:-

\begin{table}[ht]
  \centering
  \setlength{\tabcolsep}{2\tabcolsep}
  \renewcommand{\arraystretch}{1.5}
  \caption{Data Information}
  \begin{tabular}{l|l|l|l}
    \hline
\textbf{S.No} & \textbf{Column} & \textbf{Count} & \textbf{Type} \\\hline
 0 &  city    &    19524 non-null &  object\\
 1 &  name    &    19524 non-null &  object\\
 2 &  state   &    19524 non-null  & object\\
 3 &  country  &   19524 non-null &  object\\
 4 &  extnZip  &   19524 non-null  & object\\
 5 &  shortID  &   19524 non-null &  object\\
 6  & zipCode  &   19524 non-null &  object\\
 7  & streetName & 19524 non-null  & object\\
 8  & Address & 19524 non-null  & object\\
 \hline
  \end{tabular}
\end{table}

\subsection{\textbf{BM25}}
The BM25 weighting scheme [14-15], is a probabilistic model to calculate similarity without adding too many extra boundaries. We won't go into the whole hypothesis behind the model here; instead, we'll show you a series of forms that lead to the standard document scoring form.
\begin{equation}
 \displaystyle {RSV_{d}} = \sum_{t \epsilon q}^{}\log_{}\frac{N}{df_{t}} 
 \end{equation}
This variation acts somewhat unusually: assuming a term happens in over a large portion of the records in the assortment, This model generates a negative term weight, which is almost certainly undesirable. Be that as it may, accepting the utilization of the stop list, This doesn't happen very much, and the reward for each summand can be set to 0.

\begin{equation}
 \displaystyle {RSV_{d}} = \sum_{t \epsilon q}^{}\log_{}\ \left[\frac{N}{df_{t}}\right]\\ \cdot \frac{({k_1+1}){ {tf}_{td}}}{{k_1}(({1-b})+b\times(\frac{L_d}{L_ave}))+tf_{td}} 
\end{equation}

Here, The length of document \textbf{d} and the standard archive length for the entire assortment are \textbf{L\textsubscript{d}} and \textbf{L\textsubscript{ave}}, respectively, and \textbf{tf\textsubscript{td}} is the frequency of term t in document d.  A \textbf{k\textsubscript{1}} estimation of 0 compares to a paired the prototype(no term recurrence), and huge worth relates to utilizing crude term recurrence. \textbf{b} is a different tuning boundary \textbf{(0 $\leq$ b $\leq$ 1 )} that decides the scaling according to the duration of the records: \textbf{b=1} relates to completely dividing the term weight by the period of the article, while \textbf{b=0} when compared to the absence of length standardisation. In event that the question is long, we may likewise utilize comparable balancing for inquiry conditions. This is suitable if questions are passage has long data requirements, yet pointless for brief inquiries.
\begin{equation}
\begin{split}
 \displaystyle {RSV_{d}} =  \sum_{t \epsilon q}^{}\ log_{}\ \left[\frac{N}{df_{t}}\right] \cdot \frac{({k_1+1}){{tf}_{td}}}{{k_1}(({1-b})+b\times(\frac{L_d}{L_ave}))+tf_{td}} \\ 
 \times \frac{({k_3 +1}){tf}_{tq}}{k_3+{tf}_{tq}} 
\end{split}
\end{equation}
With \textbf{tf\textsubscript{tq}} denoting the frequency of term \textbf{t} in analysis \textbf{q}, and \textbf{k\textsubscript{3}} denoting yet another constructive tuning parameter, this time for the query's term frequency scaling. There is no duration normalisation of queries in the equation given (it is as if \textbf{b=0} here). Length standardizing the question is pointless on the grounds that recovery is being finished as for a solitary fixed inquiry. The tuning boundaries of these recipes ought to in a perfect world be set to upgrade execution on an improvement test assortment [16].  That is, we can look for estimations of these boundaries that boost execution on a different improvement test assortment (either physically or with streamlining strategies, for example, lattice search or something further developed), and afterward utilize these boundaries on the real test assortment. In the absence of such optimization, studies have shown that setting \textbf{k\textsubscript{1}} and \textbf{k\textsubscript{3}} to a value between 1.2 and 2 and \textbf{b=0.75} are rational values.
If relevance judgments are available, the full form of smoothed-r\textsubscript{f} can be used instead of the approximation \textbf{log(N / df\textsubscript{t})} implemented in \textbf{prob - id\textsubscript{f}}:

\begin{equation}
{
\scalebox{0.8}{$
    \begin{aligned}
     \displaystyle {RSV_{d}} \& = 
      \sum_{t \epsilon q}^{}\log_{} \left[\frac{(|{VR}_{t}+\frac{1}{2})/({VNR}_{t}+\frac{1}{2})}{(df_{t}-|{VR}_{t}|+\frac{1}{2})/(N-{df}_{t}-|{VR}|+|{VR}_{t}|+\frac{1}{2})} \right] \\ 
      \times \frac{({k_1+1}){ {tf}_{td}}}{{k_1}(({1-b})+b
     \times(\frac{L_d}{L_ave}))+tf_{td}} \times \frac{({k_3 +1}){tf}_{tq}}{k_3+{tf}_{tq}}
    \end{aligned} 
    $}
}
\end{equation}
Here, \textbf{VR\textsubscript{t}} , \textbf{NVR\textsubscript{t}} , and \textbf{VR} are used in this reference. The initial segment of the articulation reflects pertinence input (or just \textbf{id\textsubscript{f}} weighting if no pertinent data is accessible), The third considers term recurrence in the query, and the resulting carry-out reports term recurrence and study duration scaling.
As opposed to simply giving a term weighting technique to terms in a client's inquiry, pertinence input can likewise include enlarging the question (consequently or with a manual audit) with a few (say, ten to twenty) of the most important words in the referred to applicable reports as requested by the importance factor \textbf{$\vec{c}$}, and the above recipe would then be able to be utilized with a particularly expanded question vector. \textbf{$\vec{q}$} 
\newline
The BM25 measurement formulas have been working successfully across a good number of datasets and search assignments.

\subsection{\textbf{Results with BM25}}
Let’s look at some of the results when we pass a query to BM25 algorithm. In TABLE II ,III and IV we have shown the top 10 matches for different queries.
\begin{table*}
  \centering
  \caption{Query 1: MIDWEST MANUFACTURING/MW PREHUNG PLANT – BLDG \#320,14320 COUNTY ROAD 15, HOLIDAY CITY, OH 43554,TEL: 419-485-6584,ATTN: JUSTIN CRAWFORD eMAIL: jcrawfor@midwestmanufacturing.com.
 }
  \renewcommand{\arraystretch}{1.5}
  \begin{tabular}{l|c|r}
    \hline
 RANK & SCORE BY BM25 & ADDRESSES\\
 \hline
 1.& 13.774420750916285	& MIDWEST PREHUNG - PLANO, 2616 ELDAMAIN ROAD BLDG 220, PLANO, IL, 60545, USA\\
2. &	12.889157018210865 &	MIDWEST MANUFACTURING, 14319 COUNTY ROAD 15, HOLIDAYCITY, OH, 43554, USA\\
3. &	11.996978844561882 &	MIDWEST MANUFACTURING (MENARD INC.), 14320 COUNTY ROAD 15, HOLIDAYCITY, OH, 43554, USA\\
4. &	11.783313402792523 &	MWD LOGISTICS, 7236 JUSTIN WAY, MENTOR, OH, 44060\\
5. &	11.591774097246052 &	MIDWEST LIFTS LLC., 9425 COUNTY ROAD 101 N, CORCORAN, MN, 55340\\
6. &	10.836622386647267 &	AVERY, 7236 JUSTIN WAY, MENTOR, OH, 44060, USA\\
7. &	10.388212705043992 &	MIDWEST FIREWORKS WHOLESALERS LLC, 13800 COUNTY ROAD 30, BLAIR, NE, 68008, USA\\
8. &	8.256204999008487 &	PROVIA, 1550 COUNTY ROAD 140, SUGARCREEK, OH, 44681, USA\\
9. &	8.256204999008487 &	PRO TEC, 5000 COUNTY RD, LEIPSIC, OH, 45856, USA\\
10. &	8.256204999008487 &	MENARDS, 14502 COUNTY ROAD 15, HOLIDAYCITY, OH, 43554, USA\\
 \hline
  \end{tabular}
\end{table*}

\begin{table*}
  \centering
  \caption{Query 2: Alliance Outdoor Group Inc.4949 264th St Valley, NE, 68064, United States.
 }
 \setlength{\tabcolsep}{2.5\tabcolsep}
 \renewcommand{\arraystretch}{1.5}
  \begin{tabular}{l|c|r}
    \hline
 RANK & SCORE BY BM25 & ADDRESSES\\
 \hline
1. &	22.429769685155406 &	Alliance Outdoor Group Inc, 4949 264th St , , united states, Valley, NE, 68064, USA\\
2. &	9.227377232345084 &	EFCO ALLIANCE, 2030 NORTH ALLIANCE DRIVE, SPRINGFIELD, MO, 65803, USA\\
3. &	7.577227536509569 &	SAC Menards, 4949 264Th St, Valley, NE, 68064, USA\\
4. &	7.577227536509569 &	MENARD INC, 4949 264TH STREET, VALLEY, NE, 68064, USA\\
5. &	7.577227536509569 &	MENARD INC, 4949 264TH ST, VALLEY, NE, 68064, USA\\
6. &	7.520716317093744 &	VISTA OUTDOOR, 1055 E MAIN ST, ANOKA, MN, 55303\\
7. &	7.033934109026636 &	MENARDS VALLEY DC, 4801 N 264TH ST, VALLEY, NE, 68064, USA\\
8.&	7.033934109026636 &	MENARD, INC 3539 VADC, 4949 264TH ST, VALLEY, NE, 68064, USA\\
9. &	6.781126418191681 &	WORLD OUTDOOR EMPORIUM, 1307 GRANVILLE DRIVE, WENTZVILLE, MO, 63385, USA\\
10.&	6.781126418191681 &	OUTDOOR CONCEPTS, 110 W. REICHMUTH RD, VALLEY, NE, 68064, USA\\
 \hline
  \end{tabular}
\end{table*}

\begin{table*}
  \centering
  \caption{Query 3:  HANON C/O AEL SPAN 41775-100 ECORSE RD BELLEVILLE MI 48111 UNITED STATES.
 }
 \setlength{\tabcolsep}{2.5\tabcolsep}
 \renewcommand{\arraystretch}{1.5}
  \begin{tabular}{l|c|r}
    \hline
 RANK & SCORE BY BM25 & ADDRESSES\\
 \hline
1. &	34.68803161026147 &	AEL SPAN LLC, 41775 ECORSE ROAD SUITE 100, BELLEVILLE, MI, 48111, USA\\
2. &	21.211755995430522 &	FORD/CO/COUGHLIN, 41873 ECORSE RD, BELLEVILLE, MI, 48111, USA\\
3. &	21.211755995430522 &	EXEL, 41873 Ecorse Road, BELLEVILLE, MI, 48111, USA\\
4. &	20.701693493956583 &	HANON SYSTEMS, 8652 HAGGERTY ROAD DOCK 10-16, BELLEVILLE, MI, 48111, USA\\
5. &	20.392569503272448 &	AEL-SPAN AMERICA, 41775 ECORSE RD, BELLEVILLE, MI, 48111, USA\\
6. &	18.93040554854972 &	FORD MOTOR COMPANY, 41873 ECORSE ROAD SUITE290, BELLEVILLE, MI, 48111, USA\\
7. &	18.93040554854972 &	EXEL WAREHOUSE, 41873 ECORSE RD STE 290, BELLEVILLE, MI, 48111, USA\\
8. &	18.275231023758177 &	HYUNDAI GLOVIS LLC, 41873 ECORSE ROAD SUITE 200, BELLEVILLE, MI, 48111, USA\\
9. &	18.275231023758177 &	DHL SUPPLY CHAIN, 41873 ECORSE ROAD STE 250, BELLEVILLE, MI, 48111, USA\\
10. &	18.275231023758177 &	ALLIED ROSE MOVING, 41775 ECORSE RD. SUITE 190, BELLEVILLE, MI, 48111, USA\\
 \hline
  \end{tabular}
\end{table*}

On analyzing the result we can see that we were able to get most of the address correctly but there were few addresses where the best match was not given the highest score. To overcome that problem we build up an algorithm which helped us to find best match.
We saw a trend that if the difference between score of address at \textbf{1\textsuperscript{st}} position and \textbf{2\textsuperscript{nd}} position is greater than a threshold (can be find using k-means algorithm) then the address is the best match.
If the above case in not true we go for the string matching using BERT.

\subsection{\textbf{Improvements using BERT and Result}}
We utilize a language portrayal model called BERT, which represents Bidirectional Encoder Representations from Transformers.  BERT is intended to pre-train profound bidirectional portrayals from the unlabeled content by together molding on both left and right setting altogether layers. Subsequently, the pre-prepared BERT model can be calibrated with only one extra yield layer to make best-in-class models for a wide scope of undertakings, for example, question noting and language derivation, without significant errand explicit design adjustments [17]. To get the best match where Bm25 score is not greater than a particular threshold, we can use string matching by breaking address into subparts like Street Name, Company name, State , Zip code etc. Weights for each component have been shown in TABLE V.

\begin{table*}[ht]
  \centering
  \caption{Weights of each component.}
  \setlength{\tabcolsep}{10\tabcolsep}
 \renewcommand{\arraystretch}{1.5}
  \begin{tabular}{|l|r|}
    \hline
         \textbf{Component} & \textbf{Weight} \\\hline
             name & 0.21 \\\hline
             streetName & 0.23 \\\hline
             city & 0.19 \\\hline
             state &	0.16 \\\hline
             zipCode & 0.11 \\\hline
             extnZip & 0.0 \\\hline
             country & 0.1 \\
         \hline
  \end{tabular}
\end{table*}

After applying the above weights and use string similarity we are able to get the score of top 3 addresses and on the basis of score we’ll able to even categorize address into highly matched, medium match and Low match.
Results with BM25 and BERT have been shown in TABLE VI.

\begin{table*}[ht]
  \centering
  \caption{RESULTS WITH BM25 and BERT}
  \setlength{\tabcolsep}{\tabcolsep}
 \renewcommand{\arraystretch}{1.2}
 \hspace*{-0.2in}
  \begin{tabular}{|l|}
    \hline
Query 1: MIDWEST MANUFACTURING/MW PREHUNG PLANT – BLDG \#320,14320 COUNTY ROAD 15, HOLIDAY CITY, OH 43554,\\TEL: 419-485-6584,ATTN: JUSTIN CRAWFORD eMAIL: jcrawfor@midwestmanufacturing.com\\
Category: Medium\\
MIDWEST MANUFACTURING (MENARD INC.), 14320 COUNTY ROAD 15, HOLIDAYCITY, OH, 43554, USA\\
\hline
Query 2: Alliance Outdoor Group Inc.4949 264th StValley, NE, 68064, United States\\
Category: Very High\\
Alliance Outdoor Group Inc, 4949 264th St , , united states, Valley, NE, 68064, USA\\\hline
Query 3: Admiral Moving \& Logistics, C/O Corrigan Logistics Graduate Fayetteville, 1245 E Henri De Tonti Blvd Springdale, AR 72762 -Ctc: Colton Gregory\\
Category: High\\
CORRIGAN LOGISTICS A/C ADMIRAL MOVING AND LOGISTIC, 1245 EAST HENRI DE TONTI BLVD, SPRINGDALE, AR, 72762, USA\\\hline
Query 4: QUIN GLOBAL US, INC.  5710 F STREET OMAHA, NE 68117 TEL :402-731-3636 Hours: M-F 8:00 AM to 3:00 PM\\
Category: Medium\\
QUIN GLOBAL US, 5710 F STREET, OMAHA, NE, 68117, USA\\\hline
Query 5: Chainworks, Inc. 3255 Hart Rd Jackson, MI 49201\\
Category: Very High\\
CHAINWORKS, 3255 HART ROAD, JACKSON, MI, 49201, USA\\\hline
Query 6: MID-STATE WAREHOUSE 888 O Neill Drive Hebron Oh, 43025 Justin Carder PH: 740-929-5130\\
Category: High\\
MIDSTATES WAREHOUSE, 888 O NEILL DRIVE, HEBRON, OH, 43025, USA\\\hline
Query 7: NIPPON EXPRESS HEMLOCK WH 10401 HARRISON RD STE 101 ROMULUS, MI 48174 TIM ACKER 734-740-8534\\
Category: Very High\\
NIPPON EXPRESS HEMLOCK WH, 10401 HARRISON RD, ROMULUS, MI, 48174, USA\\\hline
Query 8: HANON C/O AEL SPAN 41775-100 ECORSE RD BELLEVILLE MI 48111 UNITED STATES\\
Category: High\\
AEL SPAN LLC, 41775 ECORSE ROAD SUITE 100, BELLEVILLE, MI, 48111, USA\\\hline
Query 9: HOG SLAT INC 1112 20TH STREET NORTH HUMBOLDT, IA 50548, US\\
Category: Very High\\
HOG SLAT INC, 1112 20TH STREET NORTH, HUMBOLDT, IA, 50548, USA\\\hline
Query 10: Pivot International, Inc. C/O Election Systems \& Software, In 11208 John Galt Blvd Omaha NE 68137\\
Category: High\\
Election Systems \& Software, Inc, 11208 John Galt Blvd, Omaha, NE, 68137, USA\\\hline
  \end{tabular}
\end{table*}

\section{CONCLUSION}
Address matching is a critical errand in different area based organizations like takeout administrations and expedited service. We suggest a novel way to test semantic similarity between addresses for address matching using deep semantic address representations in this proposal.We suggest making use of BM25 and BERT to learn the semantic vector representation for a string of addresses. We further suggest using the web to gain a vast volume of context for addresses, which will greatly increase the semantic sense of addresses that could be learned. Our model has higher accuracy, precision than previously stated model as shown in TABLE VII.
\begin{table*}[ht]
  \centering
  \caption{Comparison of different Models.}
  \setlength{\tabcolsep}{3\tabcolsep}
 \renewcommand{\arraystretch}{1.3}
  \begin{tabular}{|c | c | c | c|}
    \hline
 METHODS &	PRECISION&	RECALL& 	F1 SCORE \\  
 \hline
STRING 	  &   0.6854 &	0.6232 &	0.6528 \\
DICTIONARY 	 &    0.7504& 	0.6854 &	0.7164\\
AMT 	   &  0.7752 &	0.6843& 	0.7269 \\
DEEPAM	   &  0.8249 &	0.7674 &	0.7954 \\
WORD2VEC + RF&	     0.92& 	0.89 &	0.91 \\
WORD2VEC + SVM  &  	0.87&	0.81 &	0.84\\
\hline
\textbf{BM25+BERT}     &       \textbf{1.0}&	\textbf{0.97}&	\textbf{0.97}\\
 \hline
  \end{tabular}
\end{table*}

In future studies, we want to integrate more spatial information into the deep semantic address embedding.

\end{document}